# Applying a random projection algorithm to optimize machine learning model for predicting peritoneal metastasis in gastric cancer patients using CT images

Seyedehnafiseh Mirniaharikandehei[1], Morteza Heidari[1], Gopichandh Danala[1], Sivaramakrishnan Lakshmivarahan[2], Bin Zheng[1]

[1] School of Electrical and Computer Engineering, University of Oklahoma, Norman, OK 73019, USA

[2] School of Computer Sciences, University of Oklahoma, Norman, OK 73019, USA.

## Abstract

**Background and Objective:** Non-invasively predicting the risk of cancer metastasis before surgery plays an essential role in determining optimal treatment methods for cancer patients (including who can benefit from neoadjuvant chemotherapy). Although developing radiomics based machine learning (ML) models has attracted broad research interest for this purpose, it often faces a challenge of how to build a highly performed and robust ML model using small and imbalanced image datasets.

**Methods:** In this study, we explore a new approach to build an optimal ML model. A retrospective dataset involving abdominal computed tomography (CT) images acquired from 159 patients diagnosed with gastric cancer is assembled. Among them, 121 cases have peritoneal metastasis (PM), while 38 cases do not have PM. A computer-aided detection (CAD) scheme is first applied to segment primary gastric tumor volumes and initially computes 315 image features. Then, two Gradient Boosting Machine (GBM) models embedded with two different feature dimensionality reduction methods, namely, the principal component analysis (PCA) and a random projection algorithm (RPA) and a synthetic minority oversampling technique, are built to predict the risk of the patients having PM. All GBM models are trained and tested using a leave-one-case-out cross-validation method.

**Results:** Results show that the GBM embedded with RPA yielded a significantly higher prediction accuracy (71.2%) than using PCA (65.2%) ($p<0.05$).

**Conclusions**: The study demonstrated that CT images of the primary gastric tumors contain discriminatory information to predict the risk of PM, and RPA is a promising method to generate optimal feature vector, improving the performance of ML models of medical images.

# 1. Introduction

Although the occurrence of gastric cancer has declined recently, it remains the third leading cause of cancer-related death worldwide [1]. While surgery remains the only curative treatment option, preoperative neoadjuvant chemotherapy (NAC) has demonstrated favorable results with increased therapeutic resection rates and improved survival [2]. Preventing the adverse effect of NAC, patients with different disease stages must be distinguished from each other [3] because, for each step of the disease, the treatment would be different [4]. Recent studies demonstrated that applying preoperative NAC for advanced gastric cancer patients with peritoneal metastasis (PM) yielded a much better clinical outcome and enhanced the overall survival rate [5-8]. Thus, accurate assessment of the presence of the PM is essential for the selection of appropriate patients for NAC. Since the overall accuracies of subjectively reading endoscopic ultrasound and computed tomography (CT) images are not completely reliable [3, 4], an alternative technique is needed to facilitate the assessment of tumor stages and the risk of PM.

Recently, many previous studies have revealed that novel radiomics technique could extract quantitative information from medical images with a large pool of image features, and the data mining of image feature pool offers an exciting approach to build machine learning (ML) models and predict clinical outcomes [9, 10]. Although several radiomics based ML models have been reported to differentiate and stage gastric cancer patients [11, 12], these studies computed radiomics features from the tumor region manually segmented from one CT slice selected by the radiologists. Meanwhile, the correlation analysis based method was used to select a small set of image features, which cannot eliminate the redundancy of the selected features. Thus, discriminatory power and prediction accuracy of these ML models were limited. To overcome such limitations, we in this study propose to develop and evaluate a new computer-aided detection

(CAD) scheme aiming to predict the risk of PM among gastric cancer patients. First, our scheme segments primary gastric tumor volume in 3D CT image data, which can better compute image features related to the heterogeneity of the tumors. Second, to reduce the dimensionality of feature space and better identify orthogonal or non-redundant image features from a large pool of initially computed radiomics features, we investigate and apply a random projection algorithm (RPA). Third, to avoid bias in generating feature vector, RPA is embedded in a multi-feature fusion-based machine learning (ML) model to predict the risk of PM, which is trained and tested using (1) a synthetic minority oversampling technique (SMOTE) to balance numbers of cases in two classes and (2) a leave-one-case-out (LOCO) cross-validation method. The details of the study design, experimental procedures, data analysis results, and discussions are presented in the following sections of this article.

## 2. Materials and methods

2.1. Image Dataset

In this study, we use a retrospective dataset of abdominal computed tomography (CT) images acquired from 159 patients diagnosed with gastric cancers, which were confirmed by histopathology examinations of endoscopic-biopsied tissues at two hospitals. Among these patients, 121 cases have PM, and 38 cases do not have PM. Each patient had an abdominal CT imaging examination during the original cancer diagnosis, which involves approximately 300-400 image slices. The primary gastric tumors are typically depicted in around 20-22 slices. Table 1 summarizes the distribution of general demographic information of these 159 patients involved in this study.

Table 1. Distribution of study cases in the selected dataset

|  | Category | Cases with PM | Cases without PM |
|---|---|---|---|
| Total Cases |  | 121 | 38 |
| Age (years old) | < 45 | 11 (6.9%) | 5 (3.1%) |
|  | 45 – 65 | 72 (45.2%) | 23 (14.4%) |
|  | > 65 | 38 (23.8%) | 10 (6.2%) |
|  | Mean ± SD | 59.49±11.97 | 59.11±8.75 |
|  | Median | 61 | 60 |
| Gender | Men | 94 (59.1%) | 28 (17.6%) |
|  | Women | 27 (16.9%) | 10 (6.2%) |

2.2. Tumor Segmentation

By recognizing the heterogeneity of tumors in the clinical images and difficulty of tumor segmentation, we modified and implemented a hybrid tumor segmentation scheme that used a dynamic programming method [13, 14] to adaptively identify growing thresholds of a multi-layer topographic region growing algorithm and initial contour in active contour algorithm. Specifically, the tumor segmentation scheme involves the following steps. First, a Weiner filter is applied to reduce image noise. Second, an initial seed is placed at the center of the tumor region of one CT slice in which the tumor has its most significant area. To reduce the inter-operator variability in choosing the initial seed and increase the robustness of the system as demonstrated in the previous study [15], a predefined window with the size of (5,5) around the initial seed is automatically created. A pixel with the minimum value inside the window is detected and selected as the first seed point. Third, to automatically determine the first threshold value for the region growing algorithm, a new predefined window size of (5,5) is created around the new seed point. Then, the scheme computes the pixel value differences between the center pixel and boundary pixels and identifies the maximum difference. Subsequently, the region growing threshold is determined as $T_1 = V_c + 0.25 \times D_{max}$, where $V_c$ is the pixel value of the center pixel and $D_{max}$ is the computed maximum pixel value difference inside the bounding window. Then, this threshold value is applied to define the first layer of region growing to segment tumor region depicting on one CT image slice.

Fourth, after determining the first layer of tumor region growth, the growing threshold of the second layer is $T_2 = T_1 + \beta C_1$ where $C_1$ is the computed contrast of the first layer, and $\beta$ is a coefficient (i.e., 0.5). This multi-layer region growing continues until the growth ratio between two adjacent layers is two times bigger than the size of the last growing layer. Last, after the region growing algorithm stops, the scheme selects the boundary contour of the last region growing layer as initial region contour. The active contour algorithm is then applied to expend or shrink the contour curve for the best fitting tumor boundary. As a result, the scheme completes the process of segmenting the tumor region from one CT slice. Figure 1 illustrated the above steps for the tumor segmentation in one CT slice.

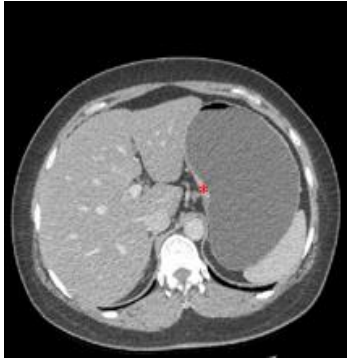
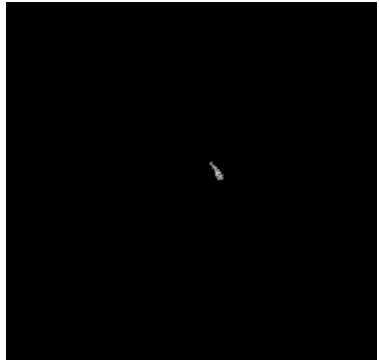
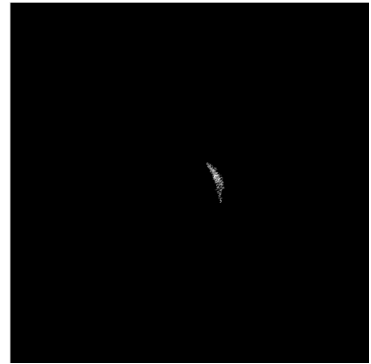

(a) Select the first seed

(b) Applying region growing based on auto initial threshold

(c) Continuing to growth while meeting the criteria of the growth

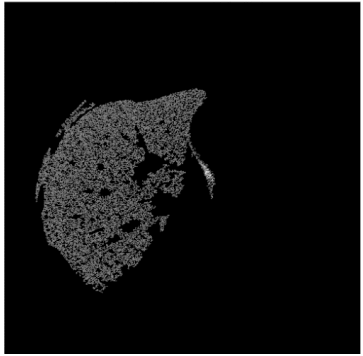
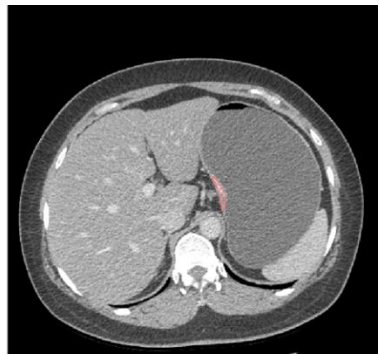

(d) Growing until the growth criterion does not meet

(e) Final segmentation

Figure 1. The process of 2D tumor region segmentation.

Subsequently, After segmenting tumor region on one CT slice, the CAD scheme continues to perform tumor region segmentation by scanning on both up and down directions until no tumor region is detected in the next adjacent CT slice. A binary map from the previous layer is obtained to perform this continuing tumor region segmentation task. Next, three seeds, including the center point of the ROI and two other points randomly selected within a predefined window around the center point, are mapped to the next slice. Then, the region growing algorithm was automatically performed from the growing seeds implemented in the targeted slice. Additionally, a tumor growing boundary condition is limited by the adjacent slice to facilitate the multi-layer region-growing process and avoid growth leakage.

Figure 2 shows an example of the segmentation of tumor regions depicting several CT image slices of one case. In this way, 3D tumor volume can be segmented and computed.

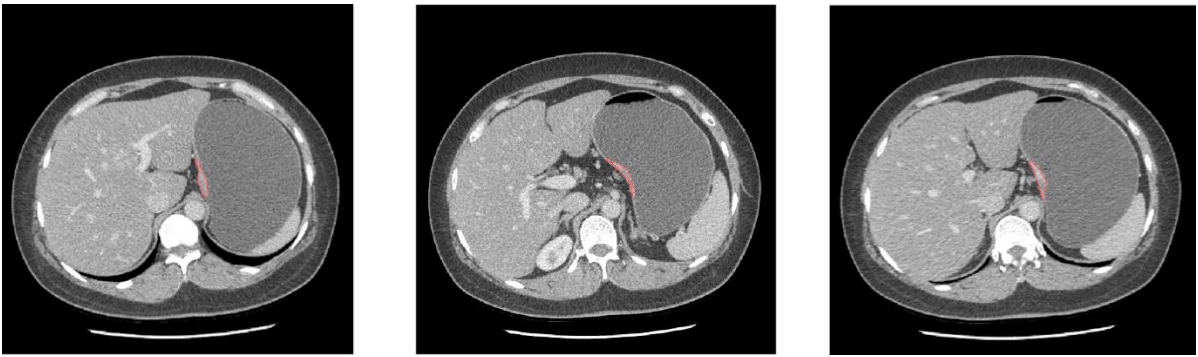

Figure 2. An Example of 3D segmentation of a lesion in 3 different slices.

2.3. Feature Extraction

Once 3D tumor volume is segmented, CAD scheme is applied to compute a large set of radiomics-based image features, which include 315 features extracted and computed from each segmented 2D tumor region (ROI) depicting on one CT image slice. These features were categorized into four main groups, including, (a) the grayscale-run length (GLRLM) features: from each ROI, 44 two dimensional features are extracted. (b) The Gray Level Difference Methods (GLDM) probability density function features: From each probability density function representing statistical texture features of ROI, four features of mean, median, standard deviation, and variance are computed. (c) Wavelet domain features: for extracting these features, first, the image is decomposed into four components comprising low and high scale decomposition in either X or Y direction by wavelet transforms. Then, the GLCM features, as well as 21 tumor density [16] and GLDM features, are extracted from those components. (d) the Laplacian of Gaussian (LoG) features: As for extracting these features first, a Gaussian smoothing filter is applied to reduce the sensitivity to the noise, and then the Laplacian filter sharpens the image's edge and highlights rapid intensity changes inside the region. Next, from the extracted points after applying the LoG filters, the mean, median, and the standard deviation are computed. Figure 3 shows the flow diagram of the feature extraction process.

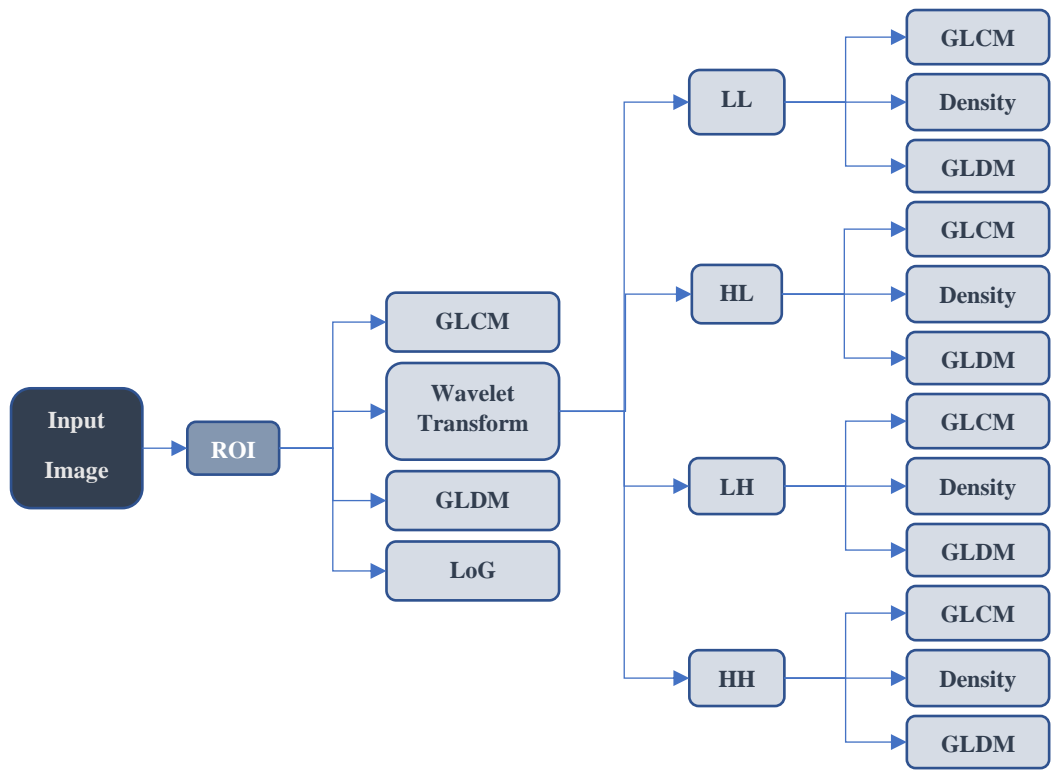

Figure 3. Diagram of Feature extraction Process.

After computing 2D features of all segmented tumor regions in $N$ involved CT image slice, CAD scheme computes each 3D feature ($F_{3D}^k$) as

$$F_{3D}^k = \sum_{i=1}^{N} w_i \times F_{2D}^k \quad (1)$$

Where $w_i$ is the ratio of the segmented tumor volume on a $i$th slice to the whole tumor volume segmented on all $N$ involved CT slices. The segmented tumor volume on a $i$ th slice is computed by multiplying the segmented region size (2D) to the CT slice thickness. Finally, all 315 computed 3D feature values are normalized between 0 to 1 to reduce case-based reliance and weight all features evenly.

2.4. Feature Dimensionality Reduction Using Random Projection Algorithm

Since the initial feature pool contains 315 image features, many of them can be redundant (highly correlated) or irrelevant (with lower performance). Hence, selecting a small set of optimal features to reduce the feature dimension and enhance learning accuracy is vital. In this study, in order to perform feature dimensionality reduction, we investigate and apply a novel image feature regeneration method of the Random Projection Algorithm (RPA). Theoretic analysis has indicated that the RPA has advantages for its simplicity, high performance, and robustness compared to other feature reduction methods; however, empirical results are sparse [17]. Meanwhile, the RPA method has been investigated and tested in many engineering applications such as text and face recognition and yielded comparable results to conventional feature regeneration methods like principal component analysis (PCA), etc.

Nevertheless, the advantage of employing RP methods over their alternative is that they generate more robust results and computationally inexpensive [17, 18]. To the best of our knowledge, the RPA has not been well investigated in the medical imaging informatics field to reduce the dimensionality of radiomics feature space. Thus, the RPA method is tested in this study.

To introduce the RPA method, let's first consider each case as a point, if the feature vector size is $k$, the case point would be in $k$ dimensional space. Thus, the Euclidian distance between two points in that k dimensional space expressed as follows:

$$|M - N| = \sqrt{\sum_{i=1}^{k}(m_i - n_i)^2} \qquad (2)$$

Regarding Formula (2), $M = (m_1, \ldots, m_k)$, and $N = (n_1, \ldots n_k)$ are two points in the k dimensional space. Likewise, the volume of a sphere with radius $r$ and volume of $V$ in $k$ dimensional space is defined as follows in Formula 3 [19]:

$$V(k) = \frac{r^k \pi^{\frac{k}{2}}}{\frac{1}{2}\Gamma(\frac{k}{2})} \qquad (3)$$

The normalization of feature matrix between [0, 1] suggests that all data can be included in a sphere with a radius of 1. The important fact about a sphere with unit radius is that the more increase in dimension, the more reduction in the volume (Formula 4). Simultaneously, the possible distance between the two points remains at 2.

$$\lim_{d \to \infty} \left( \frac{\pi^{\frac{k}{2}}}{\frac{1}{2}\Gamma(\frac{k}{2})} \right) \cong 0 \qquad (4)$$

Additionally, according to the theory of the heavy-tailed distribution, for a case like $M = (m_1, \ldots, m_k)$ in the space of features, considering features independent with an acceptable approximation, or almost perpendicular variables mapping to different axes, with $E(m_i) = p_i$, $\sum_{i=1}^{k} p_i = \mu$ and $E|(m_i - p_i)^d| \leq p_i$ for $d = 2, 3, \ldots, \lfloor t^2/6\mu \rfloor$, then, a probability can be computed using Formula 5:

$$prob(|\sum_{i=1}^{k} m_i - \mu| \geq t) \leq Max\left(3e^{\frac{-t^2}{12\mu}}, 4 \times 2^{\frac{-t}{e}}\right) \tag{5}$$

The more the value of t increases, the less chance of a point be out of that distance. Thus, $M$ should be focused around the mean value. In particular, according to Formula 4 and 5, with a satisfactory estimation, all data are contained in a sphere of unit size, and they are focused around their mean value. As a result, if the dimension increases, the volume of the sphere would close to zero. Therefore, the difference between the cases is not enough for accurate classification.

According to the above analysis, the larger the initial feature vector size, the bigger the space dimension is. Hence, most of the data is focused around the center, which leads to less difference between the features. Consequently, to reduce the feature dimension, the best and powerful technique is the one that reduces the dimensionality of features while preserves the distance between the points indicating rough preservation of the vast amount of information. If we implement a conventional feature selection method and choose a d-dimensional sup-space of the initial feature vector randomly, it is expected that all the projected distances in the new space are within a determined scale-factor of the initial k-dimensional space [20]. Thus, it is probable that after removing the redundant features, the accuracy would not increase due to the fact that the divergence between the points is not significant enough to consider as a robust model.

To address the concern discussed above and to optimize the feature space, Johnson-Lindenstrauss Lemma's theory can be applied in RPA. This theory states that for any $0 < \epsilon < 1$, and for any number of cases as $t$, which are like the points in $k$-dimensional space ($R^k$), if assuming $d$ as a positive integer, Formula 6 can be used to compute this integer number:

$$d \geq 4 \frac{\ln t}{(\frac{\epsilon^2}{2} - \frac{\epsilon^3}{3})} \tag{6}$$

Afterward, for any set $W$ of $t$ points in $R^k$, for all $z, w \in W$, it is revealed that there is a map, or random projection function like $f: R^k \to R^d$, which keeps the distance determined by Formula 7 [21]:

$$(1 - \epsilon)|z - w|^2 \leq |f(z) - f(w)|^2 \leq (1 + \epsilon)|z - w|^2 \tag{7}$$

The above approximation also can be achieved from Formula 8 as follows:

$$\frac{|f(z) - f(w)|^2}{(1 + \epsilon)} \leq |z - w|^2 \leq \frac{|f(z) - f(w)|^2}{(1 - \epsilon)} \tag{8}$$

As it is demonstrated in Formula 8, the distance between the set of points in the lower-dimension space is roughly close to the distance in high-dimensional space. The Lemma theory declares that it is feasible to project a set of points from a high-dimensional space into a lower-dimensional space, as the distances between the points are approximately preserved.

As a result, the above analysis suggests that if the initial set of features are projected into space with a lower-dimensional subspace using the random projection method, the distances between points are preserved under better contrast. Hence, it may improve the classification accuracy between the features of two classes representing cases either with or without PM under low risk of overfitting ML models. In this study, we investigate whether using RPA can yield a better result in comparison to one of the popular feature dimensionality reduction approaches, namely, principal component analysis (PCA). All extracted features in the above section are fed into both methods of RPA and PCA. After applying these two methods, each of them generates 20 optimal features out of the large initial pool of 315 features.

2.5. Machine learning model

To classify between the study cases with or without PM, we build a multi-feature fusion-based machine learning model. However, our dataset includes 121 PM cases and 38 non-PM cases, which are imbalanced in two classes. Thus, to address this issue, we apply The Synthetic Minority Oversampling Technique (SMOTE) algorithm [22] to rebalance the original image dataset. The vital point of applying SMOTE is that it introduces synthetic data by interpolation between some minority class instances that are within a specified neighborhood. If we consider $u$ as a minority class instance, it is selected as a base to generate new synthetic data points. According to a distance matrix, some nearest neighbors of the same class are chosen from the training set (points $u_1, u_4$). Then, to attain the new instances ($v_1, v_4$), a randomized interpolation is applied.

The procedure of SMOTE is as follows. Initially, the total amount of oversampling N is set up. Then, randomly, a minority class instances are chosen from the training set. Following that, the $K$ nearest neighbors are attained. Among these $K$ instances, $N$ instances are selected randomly for computing the new instances by interpolation. Figure 4 illustrates the process of creating Synthetic data in the SMOTE algorithm. As a result, we add 83 synthetic non-PM cases, and the dataset is expanded to 242 cases, including 121 PM cases and 121 non-PM cases.

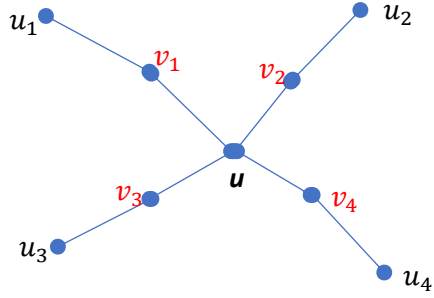

Figure 4. Synthetic data in SMOTE algorithm illustration.

After addressing the imbalance dataset, we select and implement the Gradient Boosting Machine (GBM) to train the optimal feature-based machine learning model and predict the risk of the advanced gastric cancer patients having PM. The GBM model is a popular machine learning algorithm that has proven effectiveness at classifying complex datasets and often first in class with the predictive accuracy [23]. Under a hyperparameter tuning, the GBM model is implemented to achieve a low computational cost and high robustness in detection results as well. Additionally, to decrease the case partition bias, we use a leave-one-case-out (LOCO) based cross-validation method to train and test the GBM classifier. Using the LOCO method, each case is independently tested once using the GBM model trained using all other cases in the balanced dataset. The model produces a prediction score for each testing case ranging from 0 to 1. The higher score indicates the higher risk of the test case having PM. The prediction performance is evaluated using a receiver operating characteristic (ROC) method after discarding all SMOTE generated non-PM training samples. The areas under ROC curves (AUC) and overall prediction accuracy after applying an operating threshold ($T = 0.5$) on the GBM model generated prediction scores are used as two performance evaluation indices.

Figure 5 shows the process or flow chat of using our CAD scheme to process images, compute features and train ML models in which the RPA is also embedded inside the LOCO training process to reduce potential bias of generating optimal feature vector independent to the test case. In this study, the segmentation and feature extraction steps were performed using MATLAB R2019a package, and the feature reduction and classifications were done using Python 3.7.

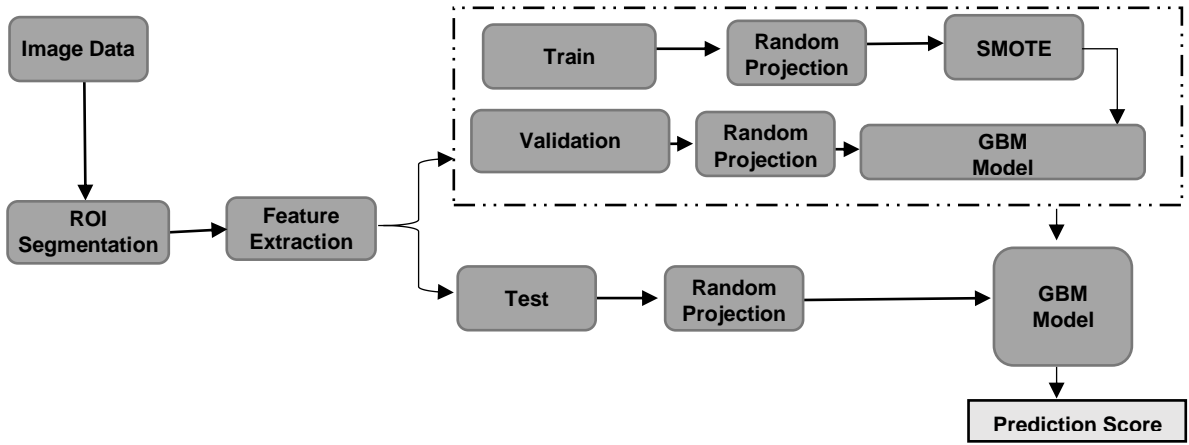

Figure 5.The flowchart of the proposed CAD scheme.

## 3. Results

Figure 6 presents two ROC curves generated by the GBM models embedded with two feature vector regeneration methods (RPA and PCA). The AUC value and the overall prediction accuracy of the GBM model trained using RPA with 3D image features as input are 0.69±0.019 and 71.2%, respectively (Figure 6.a). Further, Figure 6.b displays the AUC and accuracy of the GBM model trained using PCA, which are 0.58±0.021, and 65.2%, respectively. The results indicate that using RPA generated optimal image feature vector provides significantly higher prediction accuracy ($p < 0.05$) than using the PCA method.

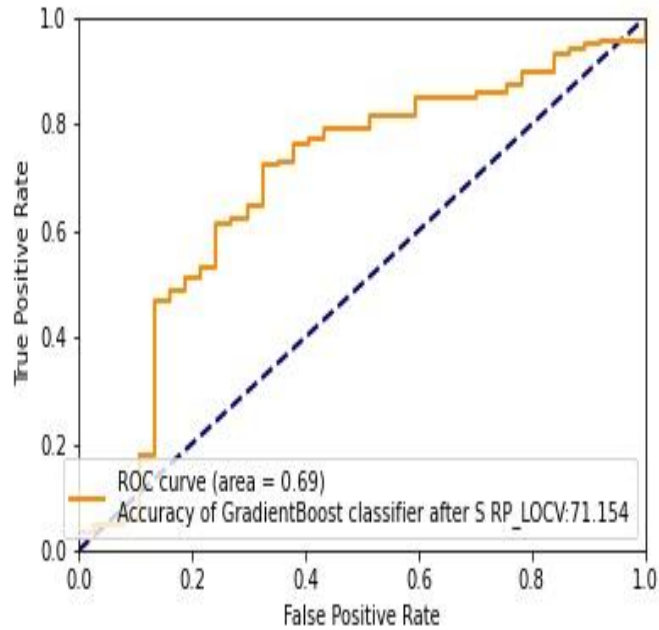

(a)

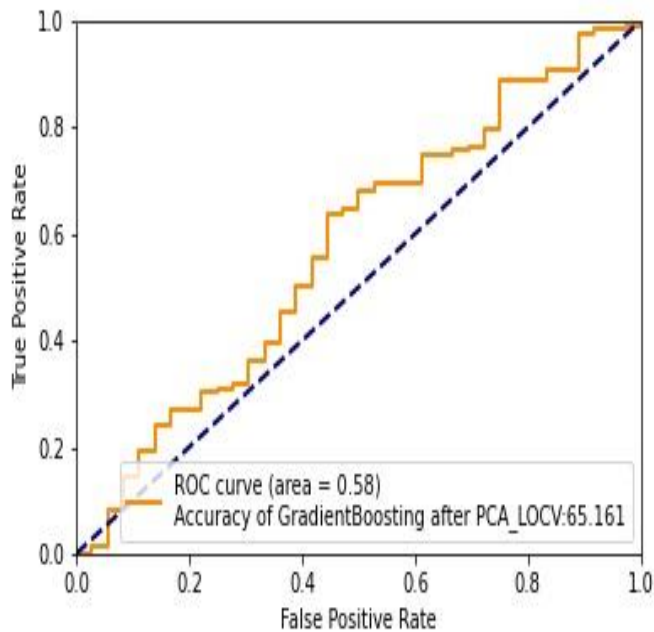

(b)

Figure 6. The ROC plot of the proposed model after applying (a)Random Projection (b)PCA for feature reduction.

Figure 7 illustrates the ROC curve and prediction performance of the GBM model trained using 2D features computed from the largest tumor region segmented from one CT image slice.

As shown in the figure, the AUC and accuracy of the GBM model are 0.66 ±0.017 and 68.4%, respectively. Table 2 shows the data to compare the performance of two GBM models built using 3D and 2D image feature vectors generated using the RPA method. The results demonstrate that using 3D image features yields significantly higher performance than using 2D features ($p < 0.05$) in predicting the risk of gastric cancer cases with PM.

Table 2. the comparison of the proposed model's performance after applying 2D and 3D features.

|  | AUC | Accuracy |
| --- | --- | --- |
| 2D features | 0.66±0.017 | 68.4% |
| 3D features | 0.69±0.019 | 71.2% |

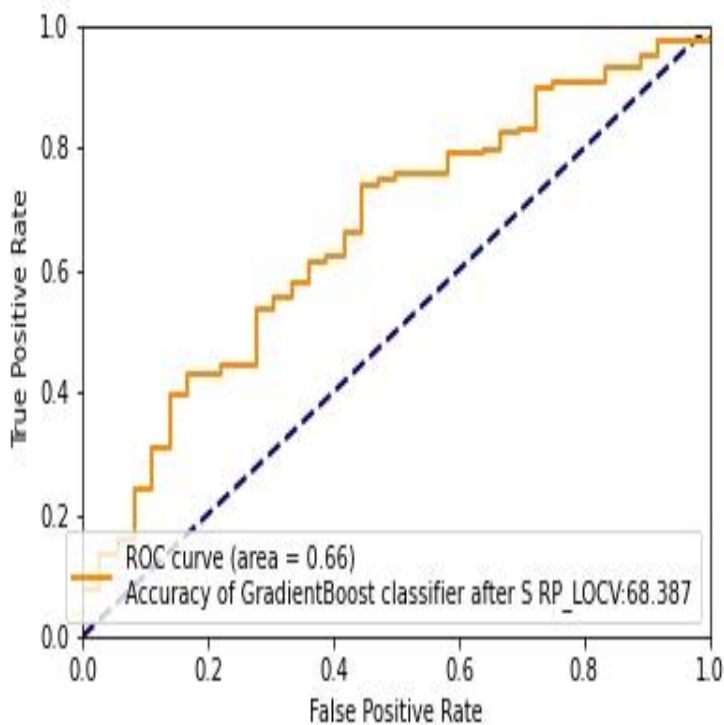

Figure 7. The ROC plot of applying the extracted 2D features to the proposed model.

## 4. Discussion

CT is the most popular imaging modality to detect and diagnose gastric cancer, and it may also provide a non-invasive alternative method to predict the risk of PM in advanced gastric cancer patients. Despite the potential advantages of using CT to detect or predict the risk of PM, the efficacy of radiologists in reading and interpreting CT images for PM detection is insufficient [24]. Thus, many studies suggested that developing and applying CAD schemes integrated with the radiomics concept and ML method can be beneficial and provide a second opinion to radiologists in detecting and diagnosing different abnormalities, including PMs of gastric cancer patients [25]. However, developing ML models using a large number of radiomics features and small training dataset remains a difficult task. In this study, we explore a new approach to develop a new CAD scheme or ML model with several unique characteristics and novel ideas in feature extraction and optimization to improve accuracy in detecting advanced gastric patients with PM.

First, in a previous study conducted in this area, the authors performed manual segmentation of gastric cancer tumor regions from the single CT image slices [26]. However, manual segmentation of tumor regions is often inconsistent with large inter-observer variability due to the fuzzy boundary of the tumor regions, which make the computed image features also inconsistent or not reproducible. Thus, the prediction accuracy can be affected or not robust. To solve this issue, we in our study developed an interactive CAD scheme with a graphical user interface (GUI) to segment tumor regions from CT images. A user only needs to place an initial seed around the center of the tumor region that has the largest size in one CT slice. Our CAD scheme then segments tumor regions on all involved CT image slices automatically. The segmentation results can also be visualized by the human eyes on the GUI windows. Although we have designed and installed correction function icon in the GUI and the user can activate this function to order CAD scheme correcting the segmentation errors (if any), the results in this study show that CAD scheme can achieve satisfactory results in automatically segmenting all 3,305 tumor regions from all 159 cases in our dataset.

Second, although several previous studies (including references [27]) have been reported to develop radiomics based ML models to detect and diagnose gastric cancer using CT images, they all used image features computed just from one manually selected CT image slice. To the best of our knowledge, this is the first study that develops and tests a new ML model using 3D image features. Our study results support our hypothesis that using 2D image features extracted from only one CT slice might not be sufficient enough to represent the heterogonous characteristics of the tumors, while using 3D image features can yield significantly higher performance. Specifically, in this study, we have performed 3D tumor segmentation and extracted 3D image features to detect or predict the risk of advanced gastric patients having PM. As shown in Table 2, the prediction performance of the GBM model trained using 3D features yield AUC=0.69±0.019 and the accuracy of 71.15%, which are significantly higher than the GBM model trained using 2D features with AUC=0.66±0.017 and the accuracy of 68.4% ($p < 0.05$), respectively.

Third, in developing CAD schemes to train machine learning classifiers, identifying a small and efficient set of image features plays a critical role; therefore, in previous studies, different feature dimensionality reduction methods have been investigated. Although these studies made many improvements in optimizing the feature vectors, there is a significant challenge of achieving small feature vectors representing the complex and non-linear image feature space. For the first time, in this study, we investigate the feasibility of applying the RPA to the medical imaging informatics field in optimizing the CAD scheme or ML model. Our study results show that RPA is a promising technique to reduce the dimensionality of a set of points lying in Euclidian space for very heterogeneous feature data commonly occurred in medical images and has advantages to achieve high robustness in classification and low risk of overfitting. Figure 6 illustrates that the classification performance of the GBM model embedded with RPA yields significantly higher performance than the GBM model embedded with a PCA, which is well-known as a popular feature dimensionality reduction method. As it was presented in Figure 6, the AUC value increased from 0.58 to 0.69 after applying the RPA as compared to the PCA. Additionally, the overall prediction accuracy of the GBM model improves from 65.2% to 71.2%, after using RPA instead of PCA. Thus, the study results demonstrate that due to very complicated distribution of radiomics features computed from medical images, RPA is a promising and more powerful technique applicable to generate optimal feature vectors for better training ML models used in CAD schemes of medical images.

Last but not least, despite the encouraging results, we also notice some limitations in this study. First, the dataset used in this study is relatively small; hence to validate the results of this study, larger datasets are required before being tested in future prospective clinical studies. Second, although in this study we have used synthetic data to balance the dataset and reduce the impact of an imbalanced dataset, using the SMOTE technique is just efficient for the low dimensional data, and it may not be appropriate or optimal for a high dimensional data [28]. Third, in the initial pool of features, we only extracted a limited number of 315 statistics and textural features, which are much less than the number of features computed based on recently developed radiomics concepts and technology in other studies [29]. Thus, more texture features can be explored in future studies to increase the diversity of the initial feature pool, which may also increase the chance of selecting or generating more optimal features to significantly improve accuracy of ML model to predict risk of PM. In summary, regardless of the limitations mentioned above, this study reveals a new and promising approach to identify and generate optimal feature vectors for training ML models implemented in the CAD schemes of medical images. Since optimizing the feature vector is one of the critical steps of building an optimal ML model, the presented method in this study is not only limited to the detection of advanced gastric patients with PM, and it can also be beneficial for other medical imaging studies of developing ML models to detect different types of cancers or abnormalities in the future.

## 5. Competing interests

The authors declare that they have no competing interests.

## 6. Authors' contributions

SM conceived of the presented idea, developed the CAD scheme and computational framework as well as analyzing the data. MH assisted with technical details in developing the CAD scheme. GD helped in carrying out the feature computation. BZH and SL supervised the project and were in charge of the overall direction. All authors provided critical feedback and helped shape the research, analysis, and manuscript.

## Acknowledgment

This study is supported in part by research grant R01 CA197150 from the National Cancer Institute. The authors also thank the support from the Stephenson Cancer Center, University of Oklahoma.